\documentclass{article}

\usepackage{PRIMEarxiv}

\usepackage[utf8]{inputenc} 
\usepackage[T1]{fontenc}    
\usepackage{hyperref}       
\usepackage{url}            
\usepackage{booktabs}       
\usepackage{amsfonts}       
\usepackage{nicefrac}       
\usepackage{microtype}      
\usepackage{lipsum}
\usepackage{fancyhdr}       
\usepackage{graphicx}       
\graphicspath{{media/}}     

\usepackage{microtype}
\usepackage{xcolor}
\usepackage{amsthm}
\newtheorem{proposition}{Proposition}
\usepackage{amsmath} 

\usepackage{natbib}
\usepackage{enumitem}
\usepackage{algorithm}
\usepackage{algorithmic}
\usepackage{multirow}
\usepackage{subcaption}

\title{Knowledge Is Not Static: Order-Aware Hypergraph RAG for Language Models
}

\author{
    Keshu Wu$^1$, Chenchen Kuai$^1$, Zihao Li$^1$, Jiwan Jiang$^1$, Shiyu Shen$^2$, \\
    \textbf{Shian Wang$^3$, Chan-Wei Hu$^1$, Zhengzhong Tu$^1$, Yang Zhou$^{1,}$\thanks{Corresponding author.}}\\
    $^1$Texas A\&M University\\
    $^2$University of Illinois Urbana-Champaign\\
    $^3$The University of Kansas\\
}

\begin{document}

\maketitle

\begin{abstract}
Retrieval-augmented generation (RAG) enhances large language models by grounding outputs in retrieved knowledge. However, existing RAG methods including graph- and hypergraph-based approaches treat retrieved evidence as an unordered set, implicitly assuming permutation invariance. This assumption is misaligned with many real-world reasoning tasks, where outcomes depend not only on which interactions occur, but also on the order in which they unfold. We propose \emph{Order-Aware Knowledge Hypergraph RAG (OKH-RAG)}, which treats order as a first-class structural property. OKH-RAG represents knowledge as higher-order interactions within a hypergraph augmented with precedence structure, and reformulates retrieval as sequence inference over hyperedges. Instead of selecting independent facts, it recovers coherent interaction trajectories that reflect underlying reasoning processes. A learned transition model infers precedence directly from data without requiring explicit temporal supervision. We evaluate OKH-RAG on order-sensitive question answering and explanation tasks, including tropical cyclone and port operation scenarios. OKH-RAG consistently outperforms permutation-invariant baselines, and ablations show that these gains arise specifically from modeling interaction order. These results highlight a key limitation of set-based retrieval: effective reasoning requires not only retrieving relevant evidence, but organizing it into structured sequences.
\end{abstract}


\medskip

\section{Introduction}
Large language models (LLMs) have demonstrated impressive capabilities in reasoning and generation, yet their performance on knowledge-intensive tasks remains constrained by the lack of explicit mechanisms for structured inference~\citep{zhao2023survey, chang2024survey, pan2024unifying, wu2025v2x}. Retrieval-augmented generation (RAG) mitigates this limitation by conditioning generation on externally retrieved knowledge, thereby separating knowledge access from language modeling~\citep{lewis2020retrieval, gao2023retrieval, yang2025graphusion, kuai2025knowledge}. While early RAG approaches rely on unstructured text retrieval, recent work has shown that incorporating \emph{structured knowledge representations} can significantly improve reasoning fidelity and factual consistency~\citep{zhu2025knowledge, fan2024survey}. 

Graph-based RAG methods represent an important step in this direction by organizing knowledge as a graph of entities and relations~\citep{edge_local_2024, he2024g, wu2024medical}. However, conventional graphs encode only \emph{binary} relationships, restricting their ability to model interactions that inherently involve more than two entities. Many phenomena in real-world systems exhibit such higher-order dependencies, where outcomes emerge from the joint configuration of multiple factors rather than pairwise interactions alone. Hypergraphs naturally generalize graphs by allowing edges to connect arbitrary-sized sets of nodes, and therefore provide a principled representation for capturing \emph{higher-order graph dynamics}~\citep{fatemi2019knowledge, bretto2013hypergraph, wu2025hypergraph, wu2025ai2, Kuai2026HowIA}.

Recent work on hypergraph-based RAG demonstrates that representing knowledge with n-ary relations can reduce information loss and improve retrieval effectiveness compared to binary graph formulations~\citep{feng2025hyperrag, luo_hypergraphrag_2025, wang2026cross, hu2026cog}. Despite this advance, existing hypergraph-based RAG methods remain fundamentally \emph{static}: hyperedges encode timeless relational facts, and retrieval is performed without accounting for how interactions evolve over time. This static abstraction limits the ability of retrieval mechanisms to support reasoning about processes, causality, and delayed effects.

\begin{figure*}[!t]
    \centering
    \includegraphics[width=0.98\textwidth]{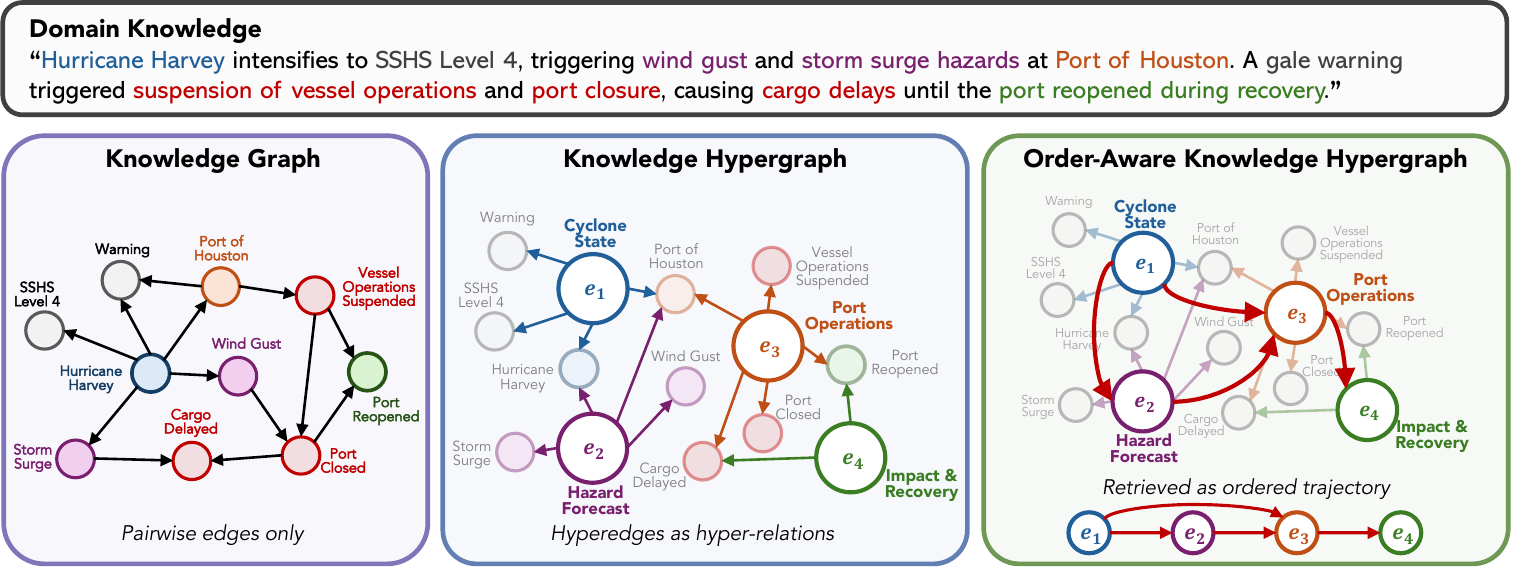}
    \caption{
    Knowledge graphs fragment multi-entity interactions into pairwise edges; knowledge hypergraphs preserve higher-order relations but retrieve them as unordered sets; our order-aware knowledge hypergraph augments hyperedges with learned precedence, enabling retrieval as ordered trajectories that reflect how interactions unfold.
    }
    \label{fig:knowledge_hypergraph}
\end{figure*}

Though many interaction-aware RAG methods have been proposed, sequential structure is fundamental to knowledge-intensive reasoning, where system-level outcomes typically emerge from ordered interactions rather than simple co-occurrences. Whether representing a chain of subsystem failures or the propagation of causal dependencies, modeling these phenomena as static relations obscures the inherent ordering, persistence, and dynamics of the system, preventing retrieval from capturing underlying dynamics. For example, in a tropical cyclone scenario, port disruption may depend jointly on forecast storm intensity, local infrastructure exposure, evolving port conditions, and operational response. This requires a hypergraph representation, since the relevant dependency is higher-order rather than pairwise. It also requires order-awareness, since the reasoning depends on how these factors evolve over time, from forecast updates to warning escalation, port restrictions, and ultimately operational disruption~\citep{kuai2025us,kuai2026cyportqa}.

In this work, we argue that \emph{order is a fundamental dimension of knowledge structure} in retrieval-augmented generation. We introduce \textbf{Order-Aware Knowledge Hypergraph RAG (OKH-RAG)}, as illustrated in Figure~\ref{fig:knowledge_hypergraph}, a framework that models knowledge as higher-order interactions with explicit precedence and reformulates retrieval as \emph{trajectory inference}. Instead of retrieving a set of independent facts, OKH-RAG retrieves \emph{ordered hyperedge trajectories} that capture how interactions evolve and connect to form coherent reasoning chains. This shifts retrieval from selecting relevant evidence to recovering structured sequences that reflect underlying processes. Our formulation unifies two complementary advances: hypergraph representations for modeling higher-order dependencies, and order-aware inference for capturing interaction dynamics. By integrating these into a single framework, OKH-RAG enables LLMs to reason over processes rather than static snapshots. Our contributions are summarized as follows:
\begin{itemize}[leftmargin=1.5em, itemsep=0pt]
    \item We introduce an \emph{order-aware knowledge hypergraph} representation that captures higher-order interactions together with their precedence structure.
    \item We reformulate retrieval as \emph{trajectory inference} over hyperedges, moving from permutation-invariant set retrieval to order-aware sequence retrieval.
    \item We integrate order-aware hypergraph retrieval into the RAG framework, enabling generation grounded in structured, temporally coherent evidence.
\end{itemize}

By integrating higher-order structure with interaction order, OKH-RAG establishes a retrieval framework where the \emph{sequence} of evidence, not merely its content, is explicitly modeled, optimized, and empirically validated as a key driver of generation quality.

\section{Related Work}


\paragraph{Retrieval-Augmented Generation.} Retrieval-augmented generation (RAG) enhances parametric language models by conditioning generation on externally retrieved knowledge, mitigating issues such as hallucination, outdated information, and lack of traceable reasoning~\citep{lewis_retrieval-augmented_2021, guu_realm_2020}. By introducing a retriever that accesses a non-parametric memory, RAG enables models to incorporate domain-specific and up-to-date information at inference time. This framework has been widely adopted in knowledge-intensive applications, particularly in domain-specific settings. Representative applications include biomedical question answering~\citep{wang_biorag_2024}, legal document analysis~\citep{pipitone_legalbench-rag_2024}, and enterprise knowledge retrieval~\citep{mishra_systematic_2025}. Early RAG systems primarily rely on retrieving unstructured text passages using dense retrieval methods such as Dense Passage Retrieval (DPR)~\citep{karpukhin_dense_2020, izacard_atlas_2022}. Subsequent surveys and benchmarks~\citep{gao_retrieval-augmented_2024, oche_systematic_2025} identify bottlenecks in both retrieval quality and scalability, particularly with the emergence of large language models (LLMs). Recent work has explored several directions to address these challenges, including knowledge-grounded dialogue systems~\citep{wang_retrieval_2024}, adaptive retrieval strategies~\citep{asai_self-rag_2023, zhai_self-adaptive_2024, yang_knowing_2025}, and multi-hop reasoning frameworks~\citep{es_ragas_2023, tang_multihop-rag_2024}. These developments collectively motivate the incorporation of structured representations into the retrieval process.

\paragraph{Graph and Hypergraph-Based RAG.}
Graph-based RAG methods~\citep{peng_graph_2024, hu_grag_2024, zhu_knowledge_2025}, which organize knowledge as entities and binary relations, represent one of the most effective extensions beyond unstructured retrieval in RAG. These approaches enable graph traversal~\citep{ma_think--graph_2024, han_retrieval-augmented_2024}, path-based reasoning~\citep{wang_reasoning_2024, chen_pathrag_2025}, and community-level summarization~\citep{edge_local_2024}, thereby improving evidence aggregation and interpretability compared to purely similarity-based retrieval. However, they remain fundamentally limited by binary representations that decompose higher-order dependencies into pairwise edges, potentially leading to information loss and fragmented reasoning. To address this limitation, hypergraph-based approaches have been introduced to connect arbitrary-sized sets of nodes, naturally capturing $n$-ary relational facts~\citep{luo_text2nkg_2024} and higher-order interactions~\citep{luo_hypergraphrag_2025, zhou_improving_2025}. Such representations provide a more expressive framework for modeling complex relationships and have shown improvements in retrieval efficiency and generation quality over binary graph structures~\citep{mavromatis_gnn-rag_2024}. Nevertheless, existing hypergraph-based RAG methods typically treat hyperedges as \emph{static} facts, failing to capture the dynamics, temporal evolution, and ordering of higher-order interactions, which limits their ability to support more complex reasoning processes.

\paragraph{Temporal Dynamics and Event-Centric Modeling.}
Temporal and dynamic graph models have been extensively studied to capture time-varying relations and node attributes~\citep{rossi_temporal_2020, cheng_temporal_2024, yang2025beyond, sun2025dyg}. These approaches typically focus on predictive tasks such as link forecasting or anomaly detection using message-passing or memory modules~\citep{galkin_message_2020, neuhauser_learning_2024}. While powerful, these models are largely designed for binary relational structures and do not directly capture higher-order interactions. Furthermore, while recent extensions~\citep{PhysRevE.109.054306} have introduced temporal hypergraphs to model evolving $n$-ary relations, they focus on representation learning rather than retrieval-augmented generation. Related work in event-centric modeling emphasizes temporal ordering and causal structure~\citep{guan_what_2023, stawarczyk_event_2021}, yet these systems often rely on task-specific schemas rather than general-purpose retrieval. OKH-RAG bridges these gaps by unifying hypergraph-based representation with order-awareness. Unlike graph-based RAG, OKH-RAG models higher-order dependencies directly via hyperedges. Unlike static hypergraph methods, it treats order as a first-class dimension, moving from set-based selection to \emph{trajectory-based retrieval}. By treating retrieval as inference over interaction sequences, OKH-RAG enables LLMs to reason about processes and propagation—capabilities previously absent in structured RAG frameworks.

\section{Methodology}
\label{sec:methodology}

\begin{figure*}[!t]
    \centering
    \includegraphics[width=1.0\textwidth]{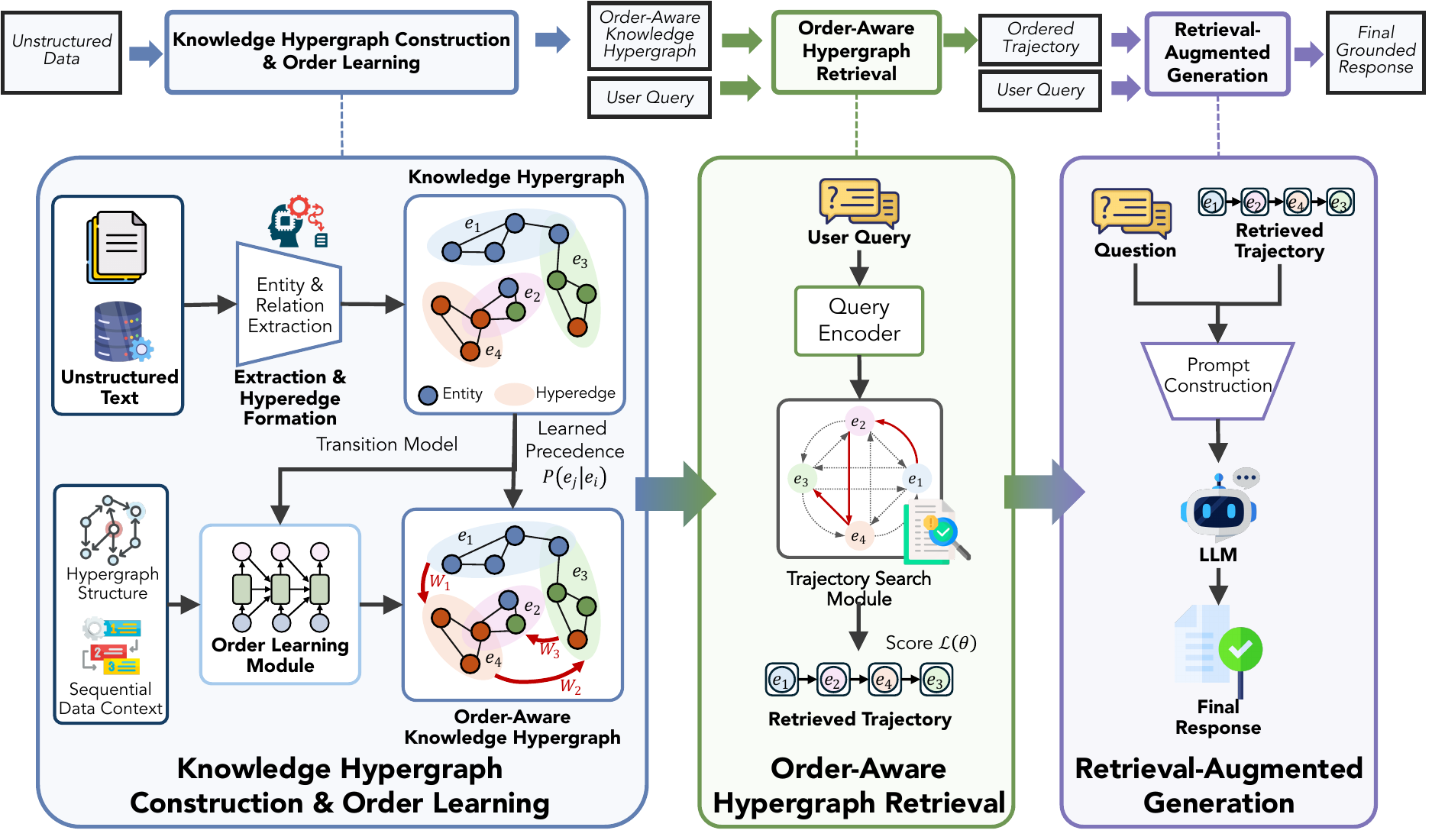}
    \vspace{-0.5em}
    \caption{
    \textbf{Overview of OKH-RAG.} The framework constructs an order-aware knowledge hypergraph from documents, retrieves query-specific interaction trajectories via sequence inference, and generates responses conditioned on structured, temporally coherent evidence.
    }
    \label{fig:okh_rag_framework}
\end{figure*}

OKH-RAG is built on the premise that the order of knowledge interactions is a structural property essential for faithful reasoning. While existing RAG systems treat retrieved evidence as an unordered set, we model knowledge as an evolving process and reformulate retrieval as trajectory inference. The framework comprises three stages (Figure~\ref{fig:okh_rag_framework}): (1) constructing an order-aware hypergraph with learned precedence (\S~\ref{sec:construction}); (2) retrieving ordered hyperedge trajectories (\S~\ref{sec:retrieval}); and (3) generating responses conditioned on structured evidence chains (\S~\ref{sec:generation}).

\subsection{Knowledge Hypergraph Construction and Order Learning}
\label{sec:construction}

The first stage transforms unstructured text into a structured, order-aware representation. Our goal is not only to capture \emph{what} interactions exist, but also \emph{how} they unfold. We first construct a knowledge hypergraph to model higher-order interactions, and then augment it with learned precedence to encode ordering.

\paragraph{Knowledge hypergraph.}
Many real-world phenomena involve interactions that are inherently higher-order: outcomes depend on the joint configuration of multiple factors rather than pairwise associations. Standard knowledge graphs, restricted to binary edges, cannot represent such interactions without fragmentation. We therefore represent domain knowledge as a \emph{knowledge hypergraph}
$\mathcal{H} = (\mathcal{V}, \mathcal{E})$,
where $\mathcal{V}$ is a set of entities and each hyperedge $e \in \mathcal{E}$ connects an arbitrary subset of entities ($|V_e| \geq 2$). Following~\citet{luo_hypergraphrag_2025}, we employ \emph{$n$-ary relational extraction} with natural language descriptions, preserving richer semantics than structured triples. While hypergraphs improve expressiveness in representing interactions, existing hypergraph-based RAG methods treat $\mathcal{H}$ as \emph{static and unordered}: retrieved hyperedges are presented as a permutation-invariant set. As we show below, this abstraction is insufficient for reasoning tasks in which order affects inference.

\paragraph{Entities.}
Each entity $v \in \mathcal{V}$ represents a distinct object, concept, or state, defined as
$v = (n_v,\; \tau_v,\; d_v,\; c_v)$,
where $n_v$ is the entity name, $\tau_v \in \mathcal{T}$ is a type from a domain-specific type system, $d_v$ is a natural-language explanation, and $c_v \in (0, 1]$ is a confidence score. The type system distinguishes \emph{persistent objects} (e.g., a port or cyclone) that anchor the hypergraph, \emph{transient states} (e.g., a cyclone state at a specific horizon) that capture evolution, and \emph{temporal anchors} (e.g., \texttt{horizon:T\nobreakdash-48}) that index interactions to positions in the ordering.

\paragraph{Hyperedges.}
Entities interact through hyperedges---the primary knowledge-carrying units. Each hyperedge is a tuple
$e = (V_e,\; r_e,\; s_e,\; \mathbf{a}_e,\; c_e)$, 
where $V_e \subseteq \mathcal{V}$ ($|V_e| \geq 2$) is the participating entity set, $r_e \in \mathcal{R}$ is a typed relation from a controlled vocabulary, $s_e$ is a \emph{natural language description} capturing the interaction's semantic content, $\mathbf{a}_e$ is a set of key--value attributes recording quantitative properties, and $c_e \in (0, 1]$ is a confidence score. The natural language description $s_e$ is critical: unlike structured triples, it can express multi-entity dependencies whose meaning arises from joint configuration rather than pairwise association. The vocabulary is defined in Appendix~\ref{app:definitions}.

\paragraph{$N$-ary relational extraction.}
We construct $\mathcal{H}$ from a document corpus $\mathcal{K}$ using a language model $\pi$ guided by an extraction prompt $p_\mathrm{ext}$:
$\mathcal{F}^d = \{f_1, \dots, f_k\} \sim \pi(\mathcal{F} \mid p_\mathrm{ext},\, d)$,
where each fact $f_i = (e_i, V_{e_i})$ pairs a hyperedge with its entity set. For multi-horizon documents, per-horizon extraction followed by merging yields higher coverage than monolithic processing. Post-extraction normalization canonicalizes identifiers, injects horizon entities, and synthesizes cross-horizon edges (Appendix~\ref{app:extraction}). The complete hypergraph aggregates across documents:
\begin{equation}
\mathcal{H} = \Bigl(\; \bigcup_{d \in \mathcal{K}} \bigcup_{f_i \in \mathcal{F}^d} V_{e_i},\;\; \bigcup_{d \in \mathcal{K}} \{e_i \mid f_i \in \mathcal{F}^d\} \;\Bigr).
\end{equation}

\paragraph{Order-aware hypergraph.}
The knowledge hypergraph captures \emph{what} interactions exist but not their order. To encode precedence while remaining agnostic to clock time, we introduce a discrete sequence index $\ell \in \{1, \dots, L\}$ capturing relative order and represent knowledge as a sequence of states $\mathcal{H}^{(\ell)} = (\mathcal{V}, \mathcal{E}^{(\ell)})$. This induces a precedence relation $e_i \prec e_j \Leftrightarrow \exists\, \ell_1 < \ell_2$ s.t.\ $e_i \in \mathcal{E}^{(\ell_1)}, e_j \in \mathcal{E}^{(\ell_2)}$, yielding the \emph{order-aware knowledge hypergraph}:
\begin{equation}
\label{eq:oakh_def}
\mathcal{H}_{\prec} = (\mathcal{V},\; \mathcal{E},\; \prec).
\end{equation}
The index $\ell$ encodes precedence, not duration; $\prec$ is a partial order accommodating concurrent interactions; and when $\prec = \emptyset$, the representation reduces to a standard unordered hypergraph. A non-empty $\prec$ breaks permutation invariance and enables order-aware retrieval. The precedence graph is constructed via domain-informed structural rules (Appendix~\ref{app:precedence}).

\paragraph{Learning precedence.}
Since explicit precedence annotations are rarely available, we learn a parametric transition model:
\begin{equation}
\label{eq:transition}
P_\theta(e_j \mid e_i)
= \frac{
    \exp(\mathbf{h}_{e_i}^\top W\, \mathbf{h}_{e_j})
}{
    \sum_{e'} \exp(\mathbf{h}_{e_i}^\top W\, \mathbf{h}_{e'})
},
\end{equation}
where $\mathbf{h}_e \in \mathbb{R}^d$ is a dense hyperedge embedding and $W$ is a learnable weight matrix (low-rank factorized as $U^\top V$, $r \ll d$). The bilinear form is inherently asymmetric---$P_\theta(e_j \mid e_i) \neq P_\theta(e_i \mid e_j)$---encoding directionality without explicit temporal features.

The model is trained via a contrastive objective with three self-supervised signals: \emph{document order} (positional adjacency as a proxy for precedence), \emph{entity-overlap consistency} (shared entities as evidence for co-participation in reasoning chains), and \emph{retrieval-induced preference} (reinforcing transitions along empirically successful trajectories in a self-training loop):
\begin{equation}
\label{eq:loss}
\mathcal{L}(\theta)
=
\mathbb{E}_{(e_i, e_j) \sim \mathcal{P}}[-\log P_\theta(e_j \mid e_i)]
+
\alpha\,\mathbb{E}_{(e_i, e_m) \sim \mathcal{N}}[\log P_\theta(e_m \mid e_i)],
\end{equation}
where $\mathcal{P}$ and $\mathcal{N}$ are positive and negative ordered pairs (details in Appendix~\ref{app:training}).

\paragraph{Order sensitivity of knowledge.}
We have described how to construct and learn $\mathcal{H}_\prec$. Is this machinery necessary? We establish formally that the answer is yes.

\begin{proposition}[Order sensitivity]
\label{prop:order_sensitivity}
There exist $k_a, k_b, q$ such that $P(y \mid k_a, k_b, q) \neq P(y \mid k_b, k_a, q)$. Any permutation-invariant retrieval method is therefore insufficient for modeling $P(y \mid q)$ when reasoning depends on interaction order.
\end{proposition}

The proposition requires no assumptions about timestamps or causality---only that evidence order can alter inferred outcomes. This complements~\citet{luo_hypergraphrag_2025}: where they establish that hypergraphs are more expressive in \emph{what} they represent, we establish that order-aware hypergraphs are more expressive in \emph{how} they can be retrieved (discussion in Appendix~\ref{app:order_sensitivity}).


\subsection{Order-aware Hypergraph Retrieval}
\label{sec:retrieval}

Standard retrieval scores each element independently and returns the highest-scoring set, treating evidence selection as a ranking problem. This discards arrangement. For order-sensitive reasoning, the quality of retrieved evidence depends not only on its members but on their sequence. We reformulate retrieval as \emph{sequence inference}: given $q$, recover the highest-scoring ordered trajectory through $\mathcal{H}_\prec$.

\paragraph{Retrieval objective.}
We seek an ordered trajectory $\gamma = (e^{(1)}, \dots, e^{(L)})$ maximizing:
\begin{eqnarray}
&~&\hskip-60pt \gamma^* = \arg\max_{\gamma}\;
\underbrace{\sum_{k=1}^{L} \mathrm{Rel}(e^{(k)}, q)}_{\text{relevance}}
+ \lambda \underbrace{\sum_{k=1}^{L-1} \log P_\theta(e^{(k+1)} \mid e^{(k)})}_{\text{order coherence}} 
\nonumber 
\\&~&\hskip100pt 
+ \mu \cdot \underbrace{\mathrm{Prec}(\gamma)}_{\substack{\text{precedence} \\ \text{consistency}}}
+ \nu \cdot \underbrace{\mathrm{Ovlp}(\gamma)}_{\substack{\text{entity} \\ \text{continuity}}}
+ \rho \cdot \underbrace{\mathrm{Cov}(\gamma)}_{\substack{\text{phase} \\ \text{coverage}}} \label{eq:retrieval_full}
\end{eqnarray}
\emph{Relevance} ensures topical pertinence via cosine similarity. \emph{Order coherence} ensures consecutive interactions form plausible transitions under $P_\theta$. \emph{Precedence consistency} enforces alignment with the structural relation $\prec$. \emph{Entity continuity} rewards entity sharing between consecutive steps, favoring coherent chains over disjointed fact assemblages. \emph{Phase coverage} rewards spanning distinct reasoning stages (advisory $\to$ hazard $\to$ operation $\to$ impact $\to$ recovery), preventing narrow evidence concentration. The hyperparameters $\lambda, \mu, \nu, \rho \geq 0$ are modular: setting all to zero recovers standard top-$k$ retrieval (formal definitions in Appendix~\ref{app:scoring}).

\paragraph{Candidate scoping and inference.}
We scope the candidate set $\mathcal{C}$ via top-$K$ cosine retrieval followed by group-aware expansion (entity overlap and domain group membership), yielding $|\mathcal{C}|$. \emph{Beam search} serves as the primary inference algorithm; \emph{Viterbi DP} provides exact optimization for smaller candidate sets (Appendix~\ref{app:inference}).

\paragraph{Multi-trajectory retrieval.}
Many queries admit multiple valid reasoning paths. We therefore retrieve a set of diverse trajectories $\Gamma_q = \{\gamma_1, \dots, \gamma_N\}$, enabling the model to consider alternative explanations and complementary evidence chains.


\subsection{Retrieval-augmented Generation}
\label{sec:generation}

Concatenating hyperedge texts into a flat context window would collapse ordered trajectories back into unordered sets. OKH-RAG instead presents each trajectory as a numbered evidence chain with explicit structural annotations.

\paragraph{Structured evidence and generation.}
Each hyperedge $e^{(k)}$ in a trajectory carries a \emph{step index}, a \emph{horizon label}, a \emph{phase label}, and \emph{entity provenance}---making the reasoning structure legible to the generator. The language model produces
$y \sim P(y \mid q, \Gamma_q)$,
and can thereby recognize preconditions, track escalation across horizons, and trace downstream consequences---capabilities inaccessible to generators receiving unordered evidence. When multiple trajectories are available, the primary mode presents all paths in a single prompt for cross-referencing; a fallback mode aggregates independent per-trajectory answers via confidence-weighted voting or averaging (Appendix~\ref{app:generation}).

\section{Experiments}
\label{sec:experiments}

We evaluate the central hypothesis of this paper: \emph{when answers depend on how evidence unfolds, retrieval should preserve order rather than treat evidence as an unordered set.} Our experiments address four questions: (1)~Does the constructed order-aware knowledge hypergraph capture the structural regimes in which order matters? (2)~Does order-aware retrieval recover coherent, query-adaptive evidence trajectories? (3)~Does this improve QA performance over permutation-invariant baselines? (4)~Do the gains arise specifically from modeling order?

\subsection{Experimental Setup}
\label{sec:setup}

We evaluate on \textbf{CyPortQA}~\citep{kuai2026cyportqa}, a domain-specific QA benchmark for tropical cyclone--port impact assessment. CyPortQA contains 2,917 real-world disruption scenarios from 2015 to 2023, spanning 145 U.S. principal ports and 90 named storms. Each scenario includes multi-horizon descriptions from T\nobreakdash-120 to T\nobreakdash-12 covering storm evolution, hazard forecasting, operational response, and impact prediction. The benchmark contains 117,178 questions across four types: True/False (TF), Multiple Choice (MC), Short Answer / Numeric (SA), and Text Description (TD). Because many questions require combining evidence across forecast horizons, CyPortQA provides a natural testbed for order-aware retrieval.

\paragraph{Baselines.}
We compare OKH-RAG against baselines that span the progression from unstructured to structured retrieval. \textbf{Text-RAG} performs dense retrieval over unstructured text chunks and concatenates the top-$k$ results as context. \textbf{GraphRAG}~\citep{edge_local_2024} retrieves over a binary knowledge graph using graph traversal and community summarization. \textbf{HyperGraphRAG}~\citep{luo_hypergraphrag_2025} retrieves unordered hyperedges from the same knowledge hypergraph $\mathcal{H}$ without modeling order. \textbf{OKH-RAG} retrieves order-aware hyperedge trajectories from $\mathcal{H}_\prec$ using the full objective in Equation~\ref{eq:retrieval_full}.
All methods use the same generator (GPT-4o) and the same embedding model (\texttt{text-embedding-3-small}), isolating the effect of retrieval structure. For OKH-RAG, we use beam search with $B=8$, $L=8$, $N=3$, and default hyperparameters $(\lambda,\mu,\nu,\rho)=(1.2,0.3,0.2,0.5)$.

\paragraph{Evaluation metrics.}
We evaluate answer accuracy across four question types. For True/False (TF) and Multiple Choice (MC), we use exact match; for Short Answer / Numeric (SA), tolerance-based accuracy; and for Text Description (TD), an LLM-judged semantic score in $[0,1]$ using GPT-4o as the evaluator.

\subsection{Structure of the constructed knowledge hypergraph}
\label{sec:retrieval_results}

\begin{figure*}[!t]
    \centering
    \begin{minipage}[t]{0.495\textwidth}
        \centering
        \includegraphics[width=\textwidth]{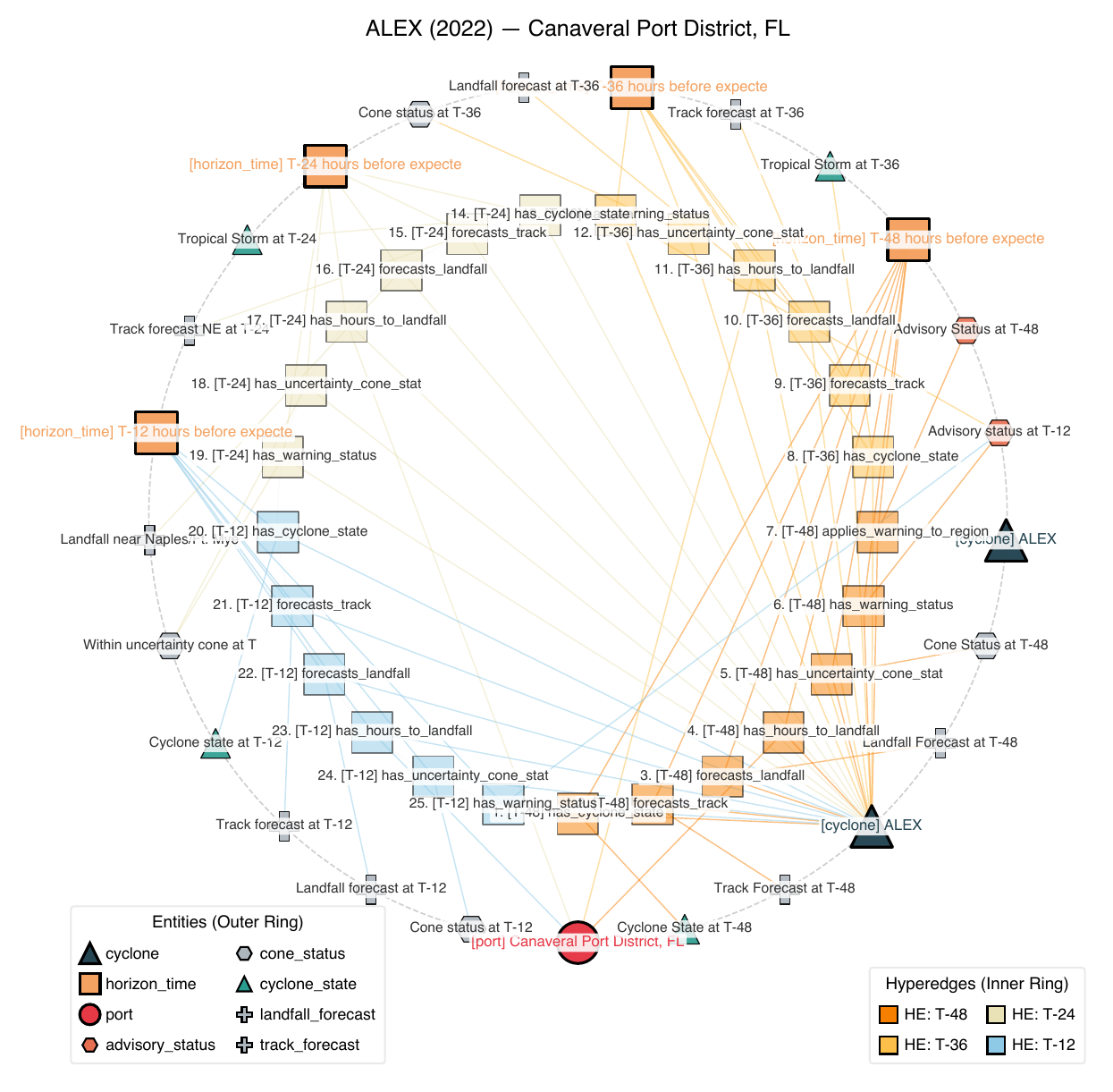}
        \subcaption{ALEX (2022), Canaveral.}
    \end{minipage}
    \hfill
    \begin{minipage}[t]{0.495\textwidth}
        \centering
        \includegraphics[width=\textwidth]{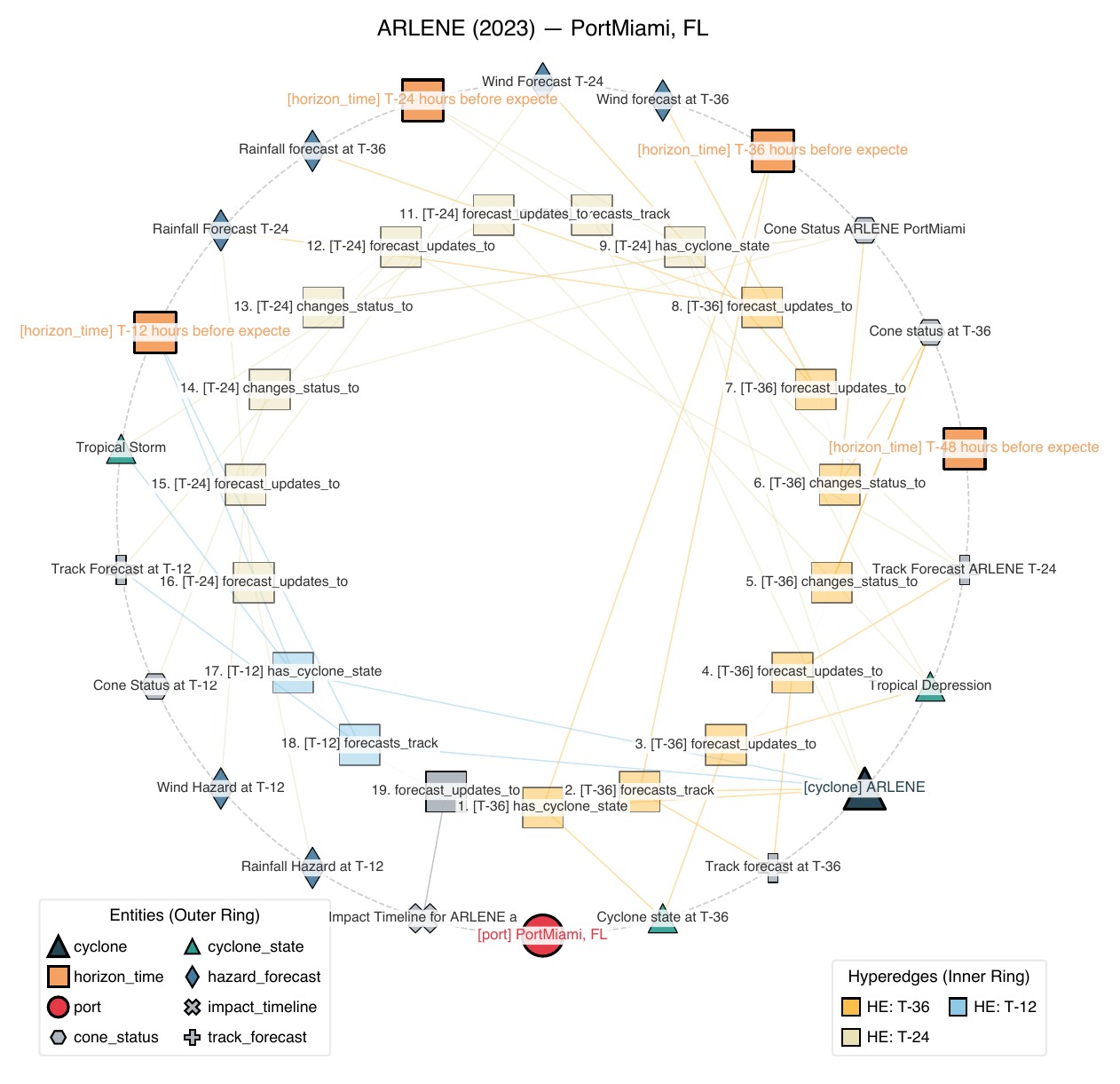}
        \subcaption{ARLENE (2023), PortMiami.}
    \end{minipage}
    \vspace{-0.5em}
    \caption{
    \textbf{Two structural regimes in the knowledge hypergraph.}
    Outer ring shows typed entities; inner ring shows ordered hyperedges.
    }
    \label{fig:radial_cases}
\end{figure*}

We first ask whether the extracted knowledge exhibits the kinds of structure that make order-aware retrieval necessary. Figure~\ref{fig:radial_cases} contrasts two scenarios at opposite ends of a structural spectrum. The hypergraph for Hurricane ALEX (left panel) shows strong \emph{within-horizon regularity}: across horizons, the extracted hyperedges follow nearly identical local phase patterns, yielding clean horizon-stratified clusters. Structural entities such as the cyclone, port, and temporal anchors connect broadly, while transient states remain largely confined to individual horizons. In this regime, unordered retrieval can often succeed by selecting the correct local snapshot. Tropical Storm ARLENE (right panel) shows a different regime. Nearly half of its hyperedges are \emph{cross-horizon transitions}, such as \texttt{forecast\_updates\_to} and \texttt{changes\_status\_to}, linking states across horizons and introducing dependencies that unfold over time. Where ALEX forms largely separable clusters, ARLENE exhibits visible inter-horizon structure. This contrast highlights when order-awareness matters most. In phase-regular scenarios, retrieval is largely local. In evolution-rich scenarios, the key challenge is not only retrieving the right hyperedges, but retrieving them in the right sequence. The order-aware hypergraph $\mathcal{H}_\prec$ is designed to support both regimes within a unified representation.

\subsection{QA retrieval results}
\label{sec:qa_retrieval}

\begin{figure*}[!t]
    \centering
    \vspace{-0.5em}
    \begin{minipage}[t]{1.0\textwidth}
        \centering
        \includegraphics[width=\textwidth, trim=0 10 0 10, clip]{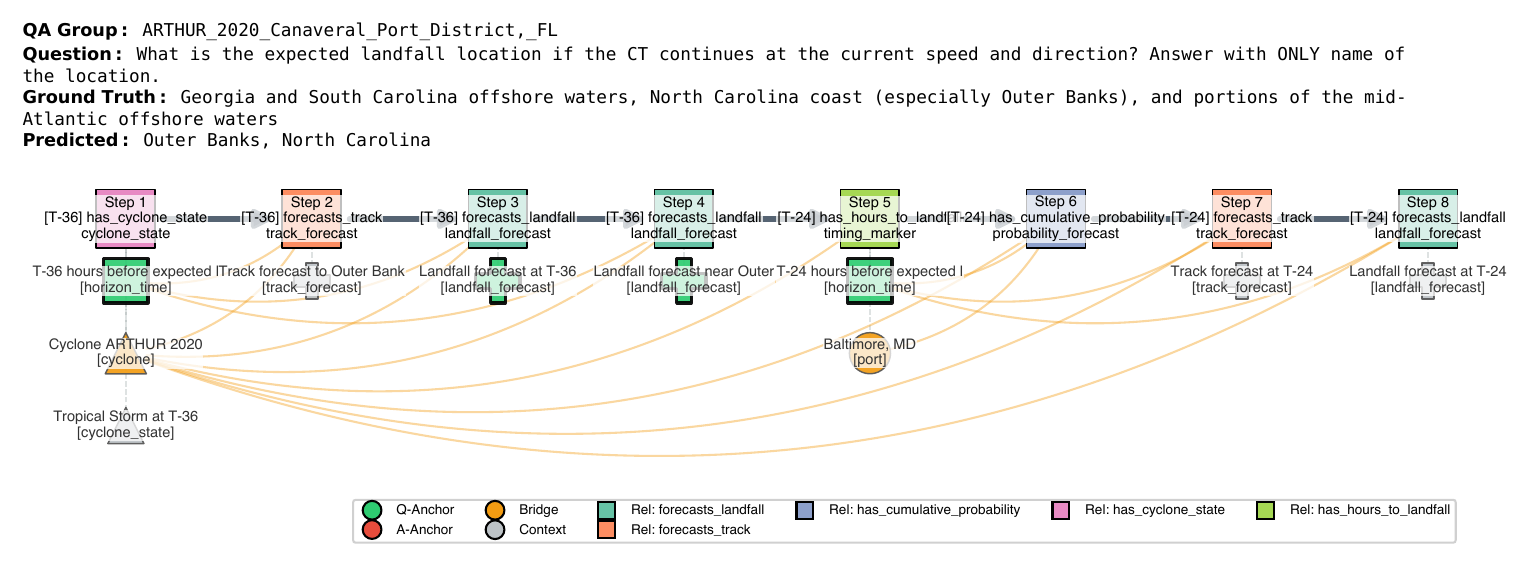}
        \subcaption{Cross-horizon reasoning.
        }
        \label{fig:traj_landfall}
    \end{minipage}
    \vspace{0.8em}
    \begin{minipage}[t]{1.0\textwidth}
        \centering
        \includegraphics[width=\textwidth, trim=0 10 0 10, clip]{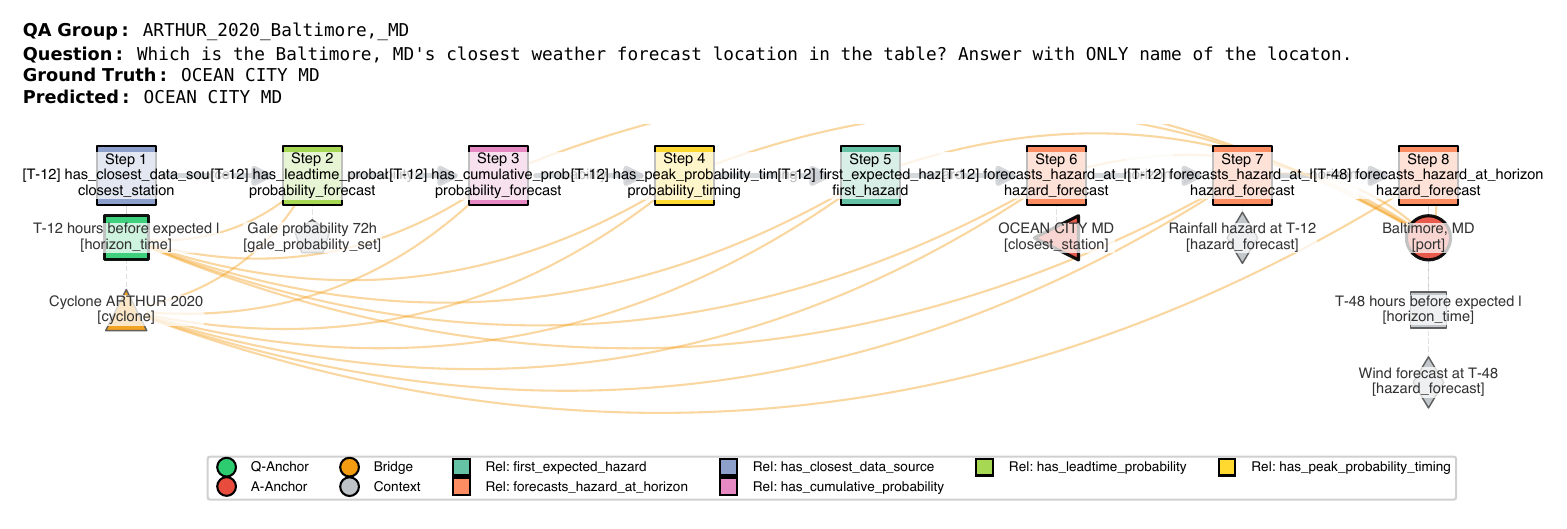}
        \subcaption{Within-horizon factual retrieval
        }
        \label{fig:traj_station}
    \end{minipage}
    \vspace{-0.9em}
    \caption{
    \textbf{Query-adaptive trajectories retrieved by OKH-RAG.}
    Upper layer: ordered hyperedges; lower layer: entities.
    }
    \label{fig:qa_trajectories}
    \vspace{-0.5em}
\end{figure*}

We next examine whether OKH-RAG adapts its retrieved trajectories to the reasoning demands of the query. Figure~\ref{fig:qa_trajectories} shows two examples for Hurricane ARTHUR (2020). The first question, ``What is the expected landfall location?'', requires \emph{cross-horizon reasoning}. The retrieved trajectory begins at T\nobreakdash-36 with cyclone state, track, and landfall evidence, then transitions to T\nobreakdash-24 for a second round of timing, probability, and landfall information. Rather than collecting isolated facts, the trajectory assembles an evolving forecast chain, allowing the generator to synthesize the answer ``Outer Banks, North Carolina'' with high confidence. The second question, ``Which is Baltimore's closest weather forecast location?'', exhibits \emph{within-horizon factual retrieval}. Here the trajectory remains at T\nobreakdash-12 and follows a compact local chain centered on the answer entity, which appears early and is reinforced later through hazard-related context. This yields an exact match with high confidence. The contrast is informative: for the landfall question, retrieval prioritizes breadth across horizons; for the station question, it prioritizes depth within a single horizon. In both cases, retrieval recovers an evidence chain aligned with the query's reasoning demands rather than a flat list of relevant hyperedges.

\subsection{Comparison with baselines}
\label{sec:baseline_comparison}

Table~\ref{tab:main_results} reports answer accuracy across all four question types. The results show a consistent progression from unstructured retrieval to structured, order-aware retrieval. Moving from Text-RAG to GraphRAG yields a large gain, indicating the value of explicit relational structure. Replacing binary graphs with hypergraphs yields a further improvement, confirming the importance of higher-order interactions. OKH-RAG performs best on all four question types and achieves the highest overall accuracy. The gain over HyperGraphRAG is particularly important because both methods use the same underlying hypergraph representation; the difference is whether retrieved evidence is treated as an unordered set or an ordered trajectory. This isolates the contribution of order-awareness beyond the representational advantage of hypergraphs alone. The largest gains appear on MC and SA, suggesting that order-aware retrieval is especially beneficial when answers require multi-step disambiguation or synthesis across related evidence.

\begin{table}[!ht]
\centering
\footnotesize
\caption{Answer accuracy by question type. Best in \textbf{bold}; second-best \underline{underlined}.}
\label{tab:main_results}
\begin{tabular}{@{}lccccc@{}}
\toprule
\textbf{Method} & \textbf{TF} & \textbf{MC} & \textbf{SA} & \textbf{TD} & \textbf{Overall} \\
\midrule
Text-RAG & 0.694 & 0.378 & 0.198 & 0.224 & 0.287 \\
GraphRAG & 0.806 & 0.506 & 0.321 & 0.362 & 0.414 \\
HyperGraphRAG & \underline{0.819} & \underline{0.620} & \underline{0.435} & \underline{0.432} & \underline{0.511} \\
OKH-RAG (ours) & \textbf{0.833} & \textbf{0.652} & \textbf{0.452} & \textbf{0.441} & \textbf{0.534} \\
\bottomrule
\end{tabular}
\end{table}

\subsection{Ablation Studies}
\label{sec:ablations}

Table~\ref{tab:ablation_master} summarizes three ablations isolating the contribution of order-aware retrieval. The permutation stress test provides the most direct evidence: \emph{OKH-RAG (shuffled)} reduces overall accuracy from 0.534 to 0.487, the largest drop observed, despite identical retrieved content. This shows that performance depends not only on which hyperedges are retrieved, but also on their order. The component analysis further indicates that all structural terms contribute positively. In particular, removing precedence consistency ($\mu$) or phase coverage ($\rho$) yields the largest degradation, suggesting that effective reasoning requires both correct ordering and sufficient coverage. Removing entity continuity ($\nu$) also reduces performance, while removing order coherence ($\lambda$) has a smaller effect, indicating its role is primarily to guide search toward better trajectories. The transition-model ablation reinforces this interpretation. Comparing \emph{no order}, \emph{heuristic order}, and \emph{learned order} reveals a consistent hierarchy: performance improves as stronger forms of order modeling are introduced. Heuristic ordering provides modest gains over order-free retrieval, while the learned transition model performs best, capturing ordering patterns beyond fixed rules. Overall, these results show that (1) evidence order is a primary driver of reasoning quality, (2) effective trajectories must be both ordered and complete, and (3) learned precedence modeling provides additional benefits beyond heuristic constraints. Together, these findings confirm that OKH-RAG’s gains arise specifically from order-aware trajectory retrieval rather than structured retrieval alone.

\begin{table*}[!ht]
\centering
\footnotesize
\caption{
\textbf{Ablation study of order-aware retrieval.}
}
\vspace{-0.8em}
\label{tab:ablation_master}
\begin{tabular}{@{}llcccccr@{}}
\toprule
\textbf{Group} & \textbf{Variant} & \textbf{TF} & \textbf{MC} & \textbf{SA} & \textbf{TD} & \textbf{Overall} & \textbf{$\Delta$} \\
\midrule
Full Model & \textbf{OKH-RAG (full)} & \textbf{0.833} & \textbf{0.652} & \textbf{0.452} & \textbf{0.441} & \textbf{0.534} & --- \\
\midrule
\multirow{1}{*}{\scriptsize Shuffle}
 & OKH-RAG (shuffled)           & 0.750 & 0.618 & 0.397 & 0.399 & 0.487 & $-$0.047 \\
\midrule
\multirow{4}{*}{\scriptsize Components}
 & $-\;\lambda$ (order coherence)   & 0.833 & 0.650 & 0.449 & 0.445 & 0.532 & $-$0.002 \\
 & $-\;\mu$ (precedence)            & 0.819 & 0.633 & 0.434 & 0.395 & 0.510 & $-$0.024 \\
 & $-\;\nu$ (entity continuity)     & 0.819 & 0.646 & 0.440 & 0.407 & 0.519 & $-$0.015 \\
 & $-\;\rho$ (phase coverage)       & 0.819 & 0.633 & 0.434 & 0.395 & 0.510 & $-$0.024 \\
\midrule
\multirow{2}{*}{\scriptsize Transition}
 & No order                     & 0.818 & 0.621 & 0.412 & 0.426 & 0.501 & $-$0.033 \\
 & Heuristic order              & 0.821 & 0.625 & 0.426 & 0.430 & 0.509 & $-$0.025 \\
\bottomrule
\end{tabular}
\end{table*}

\section{Conclusion}

This work challenges a core assumption in retrieval-augmented generation: that retrieved evidence can be treated as an unordered set. We show that this assumption fails for a broad class of reasoning tasks where outcomes depend on how interactions unfold. To address this, we introduced OKH-RAG, which represents knowledge as order-aware hypergraphs and formulates retrieval as trajectory inference over higher-order interactions. Across experiments, OKH-RAG consistently outperforms permutation-invariant baselines, and ablations confirm that these gains arise specifically from modeling interaction order rather than from structured representation alone. These results highlight that retrieval quality depends not only on relevance, but also on organization: preserving the structure of evidence sequences is critical for faithful reasoning. More broadly, our findings suggest a shift from set-based to trajectory-based retrieval, enabling LLMs to reason over processes, dependencies, and evolving systems. This perspective is particularly important for domains such as scientific discovery, decision-making under uncertainty, and complex system analysis, where order is intrinsic to meaning.

\bibliographystyle{unsrt}
\bibliography{RAG,RAG_extend}

\appendix

\section{Detailed Formulations of Methodology}
\label{app:methodology_details}

This appendix provides the formal definitions, algorithmic details, and design rationale supporting the methodology in \S~\ref{sec:methodology}. We follow the same three-stage organization: knowledge hypergraph construction (\S~\ref{app:definitions}--\ref{app:extraction}), order learning (\S~\ref{app:precedence}--\ref{app:training}), retrieval (\S~\ref{app:order_sensitivity}--\ref{app:inference}), and generation (\S~\ref{app:generation}).

\subsection{Formal definitions}
\label{app:definitions}

The main text introduces entities and hyperedges as tuples. Here we expand on the design rationale behind each component and provide the complete relation vocabulary.

\paragraph{Entity.}
An entity $v \in \mathcal{V}$ is a tuple
\begin{equation}
v = (n_v,\; \tau_v,\; d_v,\; c_v),
\end{equation}
where $n_v$ is the entity name, grounded in the source hyperedge description ($n_v \subseteq s_e$); $\tau_v \in \mathcal{T}$ is a type drawn from a domain-specific type system; $d_v$ is a natural-language explanation of the entity's role; and $c_v \in (0, 1]$ is a confidence score reflecting extraction certainty. The grounding constraint $n_v \subseteq s_e$ ensures traceability: every entity can be linked to a specific span in the source text, supporting provenance tracking and faithfulness evaluation.

The type system $\mathcal{T}$ distinguishes three functional categories, each serving a distinct role during retrieval:
\begin{itemize}[leftmargin=1.5em, itemsep=0pt]
    \item \emph{Persistent objects} (\textsc{port}, \textsc{cyclone}): domain-level entities that persist across the scenario. These appear in hyperedges at many sequence positions and serve as \emph{anchors}---their shared presence across steps is what entity continuity scoring ($\mathrm{Ovlp}$) exploits to favor coherent trajectories.
    \item \emph{Transient states} (\textsc{cyclone\_state}, \textsc{operation\_status}, \textsc{hazard\_forecast}): horizon-specific instances capturing the system's configuration at a particular point. Each is typically associated with a single horizon, encoding how persistent objects evolve---e.g., a cyclone's category at T\nobreakdash-96 versus T\nobreakdash-48.
    \item \emph{Temporal anchors} (\textsc{horizon\_time}, e.g., \texttt{horizon:T-48}): structural entities that index interactions to positions in the precedence relation. Every horizon-grounded hyperedge includes exactly one temporal anchor, enabling horizon-scoped candidate selection and order enforcement during retrieval.
\end{itemize}
This typology directly governs how reasoning chains are traced: persistent objects provide cross-horizon continuity, transient states provide within-horizon specificity, and temporal anchors provide the positional scaffolding that makes order-aware retrieval possible.

\paragraph{Hyperedge.}
A hyperedge $e \in \mathcal{E}$ is a tuple
\begin{equation}
e = (V_e,\; r_e,\; s_e,\; \mathbf{a}_e,\; c_e),
\end{equation}
with five components:
\begin{itemize}[leftmargin=1.5em, itemsep=0pt]
    \item $V_e = \{v_1, \dots, v_m\} \subseteq \mathcal{V}$: the participating entity set ($m \geq 2$). The higher-order cardinality of $V_e$ is what distinguishes hyperedges from binary edges and enables atomic representation of multi-factor interactions.
    \item $r_e \in \mathcal{R}$: a typed relation from the controlled vocabulary (Table~\ref{tab:relation_vocab}), categorizing the \emph{kind} of interaction and enabling phase-aware retrieval.
    \item $s_e$: a natural language description preserving the full semantic content of the interaction. Unlike structured triples, $s_e$ can express complex conditional dependencies involving multiple entities and quantitative thresholds simultaneously, and is directly used to compute the hyperedge embedding $\mathbf{h}_e$.
    \item $\mathbf{a}_e = \{(k_j, u_j)\}_{j=1}^{A}$: key--value attributes recording quantitative and categorical properties, enabling precise factual grounding during generation and numeric question answering.
    \item $c_e \in (0, 1]$: extraction confidence, available for downstream retrieval weighting.
\end{itemize}

\paragraph{Relation vocabulary.}
The vocabulary $\mathcal{R}$ spans the complete life cycle of domain processes, from initial state characterization through impact and recovery (Table~\ref{tab:relation_vocab}). The ordering of families 1--12 corresponds to the canonical phase progression used for within-horizon precedence (\S~\ref{app:precedence}); family~13 operates between horizons. All relation strings are normalized to this vocabulary via an alias mapping that absorbs variation in LLM-generated labels (e.g., \texttt{closes\_port} $\to$ \texttt{has\_operation\_status}).

\begin{table}[h]
\centering
\small
\caption{Relation vocabulary $\mathcal{R}$: 13 semantic families in canonical phase order.}
\label{tab:relation_vocab}
\begin{tabular}{@{}rll@{}}
\toprule
\textbf{\#} & \textbf{Family} & \textbf{Representative relations} \\
\midrule
1 & Cyclone state & \texttt{has\_cyclone\_state}, \texttt{has\_category\_state}, \texttt{has\_motion} \\
2 & Track \& landfall & \texttt{forecasts\_track}, \texttt{forecasts\_landfall} \\
3 & Timing & \texttt{has\_hours\_to\_landfall}, \texttt{has\_forecast\_window} \\
4 & Advisory & \texttt{has\_watch\_status}, \texttt{has\_warning\_status} \\
5 & Probability & \texttt{has\_leadtime\_probability}, \texttt{has\_cumulative\_probability} \\
6 & Hazard forecast & \texttt{forecasts\_hazard\_at\_horizon} \\
7 & Hazard observation & \texttt{observes\_hazard\_at\_horizon} \\
8 & Threshold & \texttt{has\_threshold\_status} \\
9 & Additional hazards & \texttt{has\_additional\_hazard} \\
10 & Operations & \texttt{has\_operation\_status}, \texttt{affects\_vessel\_handling} \\
11 & Impact & \texttt{has\_impact\_prediction}, \texttt{causes\_operational\_disruption} \\
12 & Recovery & \texttt{has\_recovery\_status}, \texttt{starts\_recovery} \\
13 & Cross-horizon & \texttt{forecast\_updates\_to}, \texttt{intensifies\_to}, \texttt{changes\_status\_to} \\
\bottomrule
\end{tabular}
\end{table}

\subsection{$N$-ary relational extraction}
\label{app:extraction}

The extraction pipeline converts unstructured documents into the knowledge hypergraph $\mathcal{H}$. Three aspects of the pipeline merit detailed description: the core extraction step, the per-horizon decomposition strategy, and the post-processing normalization.

\paragraph{Core extraction.}
For each document $d \in \mathcal{K}$, a language model $\pi$ (GPT-4o) receives an extraction prompt $p_\mathrm{ext}$ and produces $n$-ary relational facts:
\begin{equation}
\mathcal{F}^d = \{f_1, \dots, f_k\} \sim \pi(\mathcal{F} \mid p_\mathrm{ext},\, d),
\end{equation}
where each $f_i = (e_i, V_{e_i})$ pairs a hyperedge with its entity set. The prompt instructs the model to segment the input into knowledge fragments (yielding $s_e$), recognize and type all entities (yielding $V_e$ with $\tau_v$ and $d_v$), and assign relation labels from $\mathcal{R}$ with quantitative attributes. Output is constrained to a strict JSON schema matching the definitions above.

\paragraph{Per-horizon extraction.}
Multi-horizon scenario documents can span thousands of tokens. Single-call extraction risks truncation, entity conflation across horizons, and reduced attribute coverage. We therefore split each document at horizon headers (detected via regex patterns such as ``T-48 hours before expected landfall:'') and extract each block independently. This per-horizon strategy yields substantially higher entity and attribute coverage, at the cost of a subsequent merge step.

\paragraph{Post-processing.}
Five normalization steps ensure consistency across per-horizon extractions:
\begin{enumerate}[leftmargin=1.5em, itemsep=0pt]
    \item \emph{Relation normalization}: relation strings are mapped to $\mathcal{R}$ via the alias table and fuzzy matching, absorbing LLM-generated variation.
    \item \emph{Entity ID canonicalization}: identifiers are rewritten to a hierarchical convention encoding family, storm, port, and horizon (e.g., \texttt{wind\_fcst:IRMA:port\_arthur:T-48}), ensuring cross-block consistency.
    \item \emph{Horizon entity injection}: a canonical temporal anchor (\texttt{horizon:T-48}) is created for each detected horizon and added to all hyperedges at that horizon, making horizon membership structurally explicit.
    \item \emph{Cross-horizon edge synthesis}: for entity families at multiple horizons, synthetic hyperedges (\texttt{forecast\_updates\_to}, \texttt{changes\_probability\_to}) are created to represent evolution, providing the cross-horizon links that the precedence construction requires.
    \item \emph{Deduplication}: unique hyperedge IDs are computed as hashes of relation, entity set, and evidence text; exact duplicates are collapsed.
\end{enumerate}

Aggregation yields the complete hypergraph:
\begin{equation}
\mathcal{H} = \Bigl(\; \bigcup_{d \in \mathcal{K}} \bigcup_{f_i \in \mathcal{F}^d} V_{e_i},\;\; \bigcup_{d \in \mathcal{K}} \{e_i \mid f_i \in \mathcal{F}^d\} \;\Bigr).
\end{equation}

\subsection{Precedence construction}
\label{app:precedence}

The precedence relation $\prec$ over $\mathcal{E}$ is constructed from domain-informed structural rules that produce a DAG within each knowledge group.

\paragraph{Sequence-indexed representation.}
Hyperedges are assigned to sequence positions $\ell \in \{1, \dots, L\}$, yielding states $\mathcal{H}^{(\ell)} = (\mathcal{V}, \mathcal{E}^{(\ell)})$ and the precedence relation:
\begin{equation}
e_i \prec e_j \;\Longleftrightarrow\; \exists\, \ell_1 < \ell_2 \;\text{s.t.}\; e_i \in \mathcal{E}^{(\ell_1)},\; e_j \in \mathcal{E}^{(\ell_2)}.
\end{equation}

\paragraph{Structural rules.}
Four rule families determine $\prec$:

\emph{Within-horizon phase ordering.} At the same horizon, hyperedges are ordered by semantic phase following Table~\ref{tab:relation_vocab}: cyclone state $\to$ track $\to$ landfall $\to$ timing $\to$ advisory $\to$ probability $\to$ hazard $\to$ threshold $\to$ operation $\to$ impact $\to$ recovery. This reflects logical dependency: an operational decision presupposes a hazard assessment, which presupposes an advisory.

\emph{Cross-horizon family evolution.} Hyperedges in the same semantic family at different horizons follow decreasing horizon order: T\nobreakdash-96 $\prec$ T\nobreakdash-72 $\prec$ T\nobreakdash-48 $\prec$ T\nobreakdash-24 $\prec$ T\nobreakdash-12. This ensures that evolving information is presented chronologically.

\emph{Causal chain rules.} Four within-horizon inter-phase constraints encode dominant causal pathways: advisory $\prec$ hazard forecasts, hazard assessments $\prec$ operational decisions, hazard assessments $\prec$ impact predictions, and impact predictions $\prec$ recovery status.

\emph{Family-to-change ordering.} Within-horizon hyperedges in a semantic family precede any cross-horizon change hyperedges (family~13) for the same family, ensuring the ``before'' state is presented before the transition connecting it to the next horizon.

\paragraph{Canonical trajectory.}
Topological sorting the resulting DAG with deterministic tie-breaking (phase rank, family rank, lead time, text position) yields a canonical linear ordering per knowledge group. This ordering serves two purposes: it provides the ground-truth trajectory for document-order supervision in training (\S~\ref{app:training}), and it defines the structural precedence that $\mathrm{Prec}(\gamma)$ enforces during retrieval.

\subsection{Transition model training}
\label{app:training}

The learned transition model $P_\theta$ complements the structural precedence $\prec$ by capturing soft ordering preferences that rules alone cannot express---e.g., that a wind forecast is more likely to be followed by a surge forecast than by a recovery update, even when both are valid under $\prec$.

\paragraph{Embedding.}
Hyperedges are embedded as $\mathbf{h}_e \in \mathbb{R}^{1536}$ via OpenAI \texttt{text-embedding-3-small}. The input text concatenates the relation type, evidence string $s_e$, entity names and types, and key attribute values, providing both structural and semantic information.

\paragraph{Low-rank parameterization.}
The weight matrix $W = U^\top V$ with $U, V \in \mathbb{R}^{64 \times 1536}$ yields $196{,}608$ parameters versus $\sim$2.4M for the full matrix, while preserving the asymmetry essential for directional modeling. The partition function is approximated via sampled softmax with $K = 64$ global negatives per example.

\paragraph{Self-supervised signals.}
Three signal families, each exploiting a different form of implicit order, compose the training set:

\emph{Document order ($\mathcal{P}_\mathrm{doc}$).} For co-occurring hyperedges $(e_i, e_j)$ with $\pi(e_i) < \pi(e_j)$, the pair joins $\mathcal{P}$; the reversal $(e_j, e_i)$ and random cross-group pairs join $\mathcal{N}$. This is the most abundant but noisiest signal: document order is a proxy for, not a guarantee of, logical precedence.

\emph{Entity-overlap consistency ($\mathcal{P}_\mathrm{ent}$).} Pairs sharing entities ($|V_{e_i} \cap V_{e_j}| \geq 1$) within the same knowledge group join $\mathcal{P}$ if they appear in canonical trajectory order (\S~\ref{app:precedence}). This signal is sparser but higher-quality: shared entities imply topical relatedness, and the canonical order provides a principled direction.

\emph{Retrieval-induced preference ($\mathcal{P}_\mathrm{ret}$).} After initial training, retrieval traces from correct answers yield additional positive consecutive pairs, creating a self-training loop: better retrievals improve the model, which improves subsequent retrievals.

\subsection{Order sensitivity: proof and discussion}
\label{app:order_sensitivity}

Proposition~\ref{prop:order_sensitivity} asserts that permutation-invariant retrieval is insufficient for order-sensitive reasoning. We provide a constructive proof and discuss scope.

\paragraph{Constructive proof.}
Let $k_1, k_2, k_3$ be three interactions:
\begin{align}
k_1 &: \text{At T-96, the storm is a tropical depression (Category\,0).} \notag \\
k_2 &: \text{At T-48, the storm has intensified to Category\,2; uncertainty cone covers the port.} \notag \\
k_3 &: \text{At T-12, gale-force wind probability exceeds 80\%; port restricts vessel movements.} \notag
\end{align}
For the query ``Was the port restriction justified?'', the sequence $(k_1, k_2, k_3)$ presents progressive escalation supporting ``yes,'' while $(k_3, k_2, k_1)$ presents the restriction before its justification, supporting ``premature.'' Since these yield different answer distributions, $P(y \mid k_1, k_2, k_3, q) \neq P(y \mid k_3, k_2, k_1, q)$. \qed

\paragraph{Scope.}
The proposition requires only that \emph{some} query's answer depends on order---not that all queries do. This minimal requirement is satisfied in any domain with precedence-dependent outcomes: clinical diagnosis, legal reasoning, cascading engineering failures, or operational decision-making under evolving hazards.

\subsection{Retrieval scoring terms}
\label{app:scoring}

All terms in Equation~\ref{eq:retrieval_full} are designed for efficient incremental computation during beam search.

\paragraph{Relevance.}
$\mathrm{Rel}(e, q) = \cos(\mathbf{h}_e, \mathbf{q})$: cosine similarity in $\mathbb{R}^{1536}$ between hyperedge and query embeddings. Precomputed once per query for all candidates.

\paragraph{Order coherence.}
$\mathrm{Trans}(\gamma) = \sum_{k=1}^{L-1} \log P_\theta(e^{(k+1)} \mid e^{(k)})$: cumulative log-transition probability, computed incrementally from the precomputed $|\mathcal{C}| \times |\mathcal{C}|$ matrix.

\paragraph{Precedence consistency.}
\begin{equation}
\mathrm{Prec}(\gamma) = \frac{\bigl|\{k : e^{(k)} \prec e^{(k+1)}\}\bigr|}{\bigl|\{k : (e^{(k)}, e^{(k+1)}) \in (\prec \cup \succ)\}\bigr|}.
\end{equation}
Fraction of ordered consecutive pairs consistent with $\prec$; unrelated pairs are excluded from the denominator.

\paragraph{Entity continuity.}
\begin{equation}
\mathrm{Ovlp}(\gamma) = \frac{1}{L{-}1} \sum_{k=1}^{L-1} \frac{|V_{e^{(k)}} \cap V_{e^{(k+1)}}|}{|V_{e^{(k)}} \cup V_{e^{(k+1)}}|}.
\end{equation}
Mean Jaccard similarity of entity sets between consecutive steps.

\paragraph{Phase coverage.}
\begin{equation}
\mathrm{Cov}(\gamma) = \frac{\bigl|\{\phi(e^{(k)}) : k = 1, \dots, L\} \cap \mathcal{S}\bigr|}{|\mathcal{S}|},
\end{equation}
where $\phi(e)$ maps a hyperedge to its reasoning phase and $\mathcal{S} = \{\texttt{advisory}, \texttt{hazard\_forecast}, \texttt{hazard\_observation}, \texttt{operation\_status}, \texttt{impact\_prediction}, \texttt{recovery\_status}\}$.

\subsection{Inference algorithms}
\label{app:inference}

\paragraph{Beam search.}
Algorithm~\ref{alg:beam} formalises order-aware beam search. The key distinction from standard beam search is \emph{sequence-dependent scoring}: the score of appending $e'$ to beam $\beta$ depends on the last element (via $P_\theta$), the accumulated entity set (via $\mathrm{Ovlp}$), and the full prefix (via the diversity penalty). This dependency is what transforms retrieval from independent ranking into trajectory inference.

\begin{algorithm}[h]
\caption{Order-Aware Beam Search}
\label{alg:beam}
\begin{algorithmic}[1]
\REQUIRE Query $q$, candidates $\mathcal{C}$, model $P_\theta$, width $B$, length $L$, weights $\lambda, \mu, \nu, \rho$
\STATE Precompute $\mathrm{Rel}(e, q)$ and $\log P_\theta(e_j \mid e_i)$ for all $e, e_i, e_j \in \mathcal{C}$
\STATE $\mathcal{B} \gets$ top-$2B$ singletons by $\mathrm{Rel}(e, q)$
\FOR{$\ell = 2, \dots, L$}
    \STATE $\mathcal{B}' \gets \emptyset$
    \FOR{each $\beta = (e^{(1)}, \dots, e^{(\ell-1)}) \in \mathcal{B}$, each $e' \in \mathcal{C} \setminus \beta$}
        \STATE $s \gets \mathrm{Rel}(e', q) + \lambda \log P_\theta(e' \mid e^{(\ell-1)}) + \mu \cdot \mathbb{1}[e^{(\ell-1)} \prec e']$
        \STATE $s \gets s + \nu \cdot J(V_{e^{(\ell-1)}}, V_{e'}) + \rho \cdot \Delta\mathrm{Cov}(e', \beta)$
        \STATE $\mathcal{B}' \gets \mathcal{B}' \cup \{(\beta \oplus e',\; \mathrm{score}(\beta) + s)\}$
    \ENDFOR
    \STATE Apply diversity penalty; retain top-$B$ by score
\ENDFOR
\RETURN Top-$N$ trajectories with score breakdowns
\end{algorithmic}
\end{algorithm}

\paragraph{Viterbi dynamic programming.}
For exact optimisation under the two-term objective ($\mathrm{Rel} + \lambda \cdot \mathrm{Trans}$):
\begin{align}
\mathrm{dp}[0][j] &= \mathrm{Rel}(e_j, q), \\
\mathrm{dp}[\ell][j] &= \max_{i} \bigl\{\mathrm{dp}[\ell{-}1][i] + \mathrm{Rel}(e_j, q) + \lambda \log P_\theta(e_j \mid e_i)\bigr\},
\end{align}
with backpointers for trajectory recovery. Complexity: $O(L \cdot |\mathcal{C}|^2)$. Viterbi optimizes the two-term objective exactly but omits auxiliary terms; it serves primarily as a quality ceiling for beam search.

\paragraph{Candidate scoping.}
Two stages reduce the search space: (1)~top-$K$ ($K{=}80$) cosine retrieval from the global index; (2)~group-aware expansion via group membership and one-hop entity overlap, with 40\% of slots reserved for the query's own group when known. The pool is capped at $|\mathcal{C}| = 150$.

\subsection{Generation details}
\label{app:generation}

\paragraph{Evidence format.}
Each hyperedge $e^{(k)}$ in a trajectory is presented as:
\begin{verbatim}
[Step k] [T-XX] [phase=...] [family=...]
  Relation: r_e
  Evidence: s_e
  Reasoning: within_horizon, hazard_to_operation, ...
  Entities: entity1 [type]; entity2 [type]; ...
\end{verbatim}
Step indices provide ordering cues; horizon and phase labels provide temporal and logical context; reasoning tags indicate causal pathway participation; entity lists enable cross-step tracking. A trajectory-level quality summary is prepended when available.

\paragraph{Multi-trajectory synthesis.}
The \emph{single-call} mode presents all $N$ trajectories as numbered reasoning paths in one prompt. The model is instructed to read all paths before answering, treat convergence as a reliability signal, and note which trajectory supports the final answer. The \emph{fallback} mode generates per-trajectory answers and aggregates via confidence-weighted voting (categorical) or averaging (numeric). The single-call mode is preferred: it enables direct cross-referencing rather than post-hoc reconciliation.

\paragraph{Output format.}
Structured JSON with fields \texttt{answer}, \texttt{confidence} $\in [0,1]$, and \texttt{rationale}. Type-specific constraints ensure evaluation-friendly output: ``Yes''/``No'' for true/false, letter labels for multiple choice, numeric values for quantitative questions, free text for descriptions.

\end{document}